\documentclass[10pt,twocolumn,letterpaper]{article}

\usepackage{iccv}
\usepackage{times}
\usepackage{epsfig}
\usepackage{graphicx}
\usepackage{amsmath}
\usepackage{amssymb}

\usepackage{color}
\usepackage{booktabs}
\usepackage{dsfont}
\usepackage{multirow}
\usepackage{caption} 
\usepackage[ruled,vlined]{algorithm2e}
\DeclareMathOperator*{\argmax}{arg\,max}
\SetArgSty{textnormal}
\SetAlgoSkip{}
\usepackage[breaklinks=true,bookmarks=false]{hyperref}

\iccvfinalcopy 

\def\iccvPaperID{****} 

\ificcvfinal\fi

\begin{document}

\setlength{\abovedisplayskip}{4pt}
\setlength{\belowdisplayskip}{4pt}

\title{PANet: Few-Shot Image Semantic Segmentation with Prototype Alignment}

\author{Kaixin Wang \textsuperscript{1} \qquad Jun Hao Liew \textsuperscript{2} \qquad Yingtian Zou \textsuperscript{2} \qquad Daquan Zhou \textsuperscript{1} \qquad  Jiashi Feng \textsuperscript{2}   \\
\textsuperscript{1} NGS, National University of Singapore \quad \textsuperscript{2} ECE Department, National University of Singapore\\
{\tt\small \{kaixin.wang, liewjunhao\}@u.nus.edu  \quad \{elezouy, elefjia\}@nus.edu.sg \quad zhoudaquan21@gmail.com
}
}

\maketitle
\ificcvfinal\thispagestyle{empty}\fi

\begin{abstract}
   Despite the great progress made by deep CNNs in image semantic segmentation, they  typically require a large number of densely-annotated images for training and are difficult to generalize to unseen object categories. Few-shot segmentation has thus been developed to learn to perform segmentation from only a few annotated examples. In this paper, we tackle the challenging few-shot segmentation problem from a metric learning perspective and present PANet, a novel prototype alignment network to better utilize the information of the support set. Our PANet learns class-specific prototype representations from a few support images within an embedding space and then performs segmentation over the query images through matching each pixel to the learned prototypes. With non-parametric metric learning, PANet offers high-quality prototypes that are representative for each semantic class and meanwhile discriminative for different classes. Moreover, PANet introduces a prototype alignment regularization between support and query. With this, PANet fully exploits knowledge from the support and provides better generalization on few-shot segmentation. Significantly, our model achieves the mIoU score of 48.1\% and 55.7\% on PASCAL-5\textsuperscript{i} for 1-shot and 5-shot settings respectively, surpassing the state-of-the-art method by 1.8\% and 8.6\%.
\end{abstract}

\vspace{-4mm}
\section{Introduction}

Deep learning has greatly advanced the development of semantic segmentation with a number of CNN based architectures like FCN~\cite{long2015fully}, SegNet~\cite{badrinarayanan2017segnet}, DeepLab~\cite{chen2018deeplab} and PSPNet~\cite{zhao2017pyramid}. However, training these models typically requires large numbers of images with pixel-level annotations which are expensive to obtain. Semi- and weakly-supervised learning methods~\cite{wei2017object, dai2015boxsup, lin2016scribblesup, papandreou2015weakly} alleviate such requirements but still need many weakly annotated training images. Besides their hunger for training data, these models also suffer rather poor generalizability to unseen classes. To deal with the aforementioned challenges, few-shot learning, which learns new concepts from a few annotated examples, has been actively explored, mostly concentrating on image classification~\cite{vinyals2016matching, snell2017prototypical, sung2018learning, ravi2016optimization, finn2017model, garcia2018fewshot, liu2018learning, oreshkin2018tadam} and a few targeting at segmentation tasks~\cite{shaban2017one, rakelly2018few, dong2018few, zhang2018sg, dong2018few, Hu2018AttentionbasedMG}. 

\begin{figure}[t]
\begin{center}
   \includegraphics[width=\linewidth]{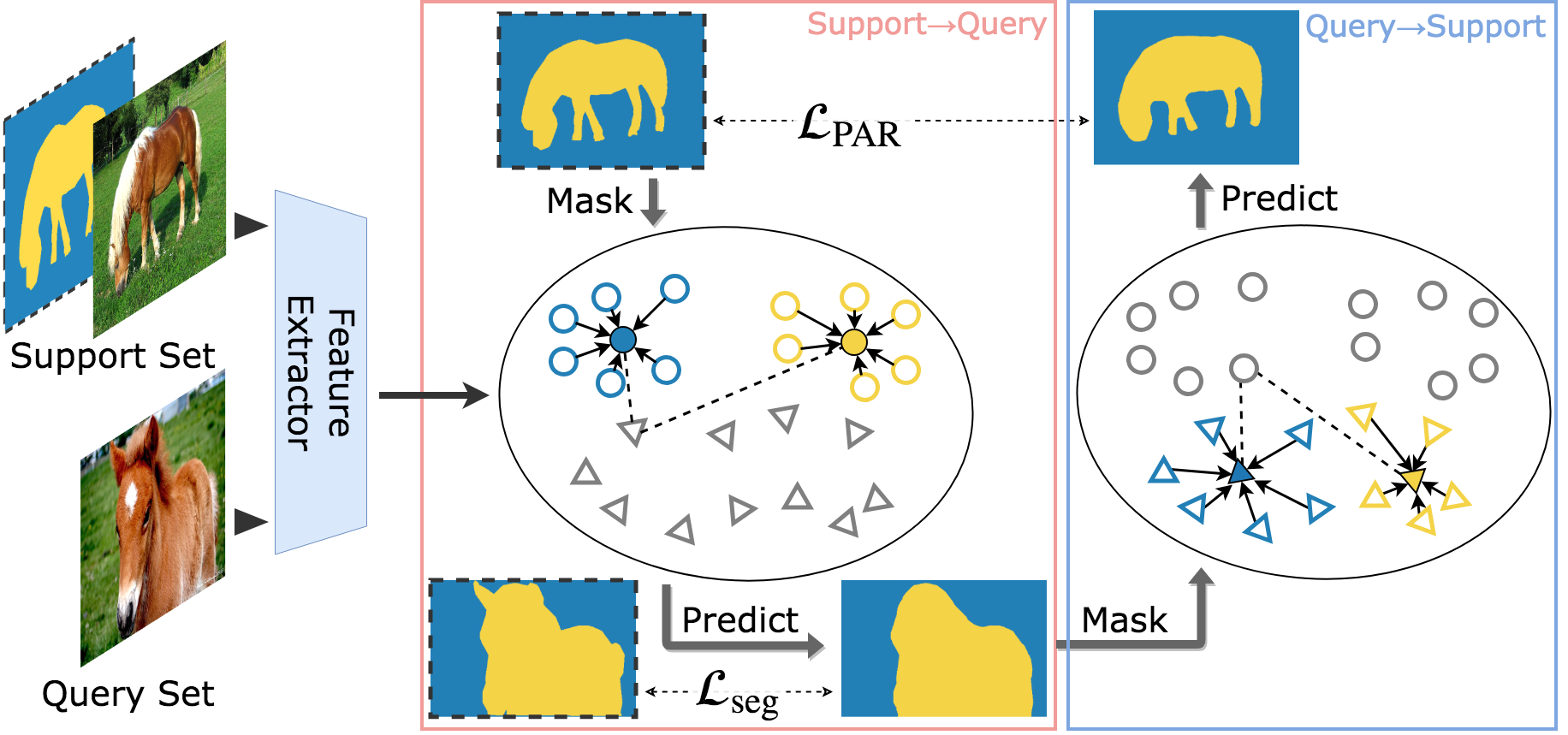}
\end{center}
   \caption{Overview of our model (PANet) for few-shot segmentation. PANet first maps the support and query images into embedding features (circles and triangles respectively) and learns prototypes for each class (blue and yellow solid circles). Segmentation over the query is then performed by matching its features to a nearest prototype within the embedding space (dashed lines). PANet further introduces a prototype alignment regularization during training to align the prototypes from support and query images within the embedding space by performing few-shot segmentation reversely from query to support (right panel). 
   Segmentation masks with dashed border denote ground truth annotations.
   }
\label{fig:overall}
\end{figure}

Existing few-shot segmentation methods generally learn from a handful of \emph{support} images and then feed learned knowledge into a parametric module for segmenting the \emph{query}. However, such schemes have two drawbacks and thus generalize unsatisfactorily. First, they do not differentiate the knowledge extraction and segmentation process, which may be problematic since the segmentation model representation is mixed with the semantic features of the support. We therefore propose to separate these two parts as prototype extraction and non-parametric metric learning. The prototypes are optimized to be compact and robust representations for each semantic class and the non-parametric metric learning performs segmentation through pixel-level matching within the embedding space. Moreover, instead of using the annotations of the support only for masking as in previous methods, we propose to leverage them also for supervising the few-shot learning process. To this end, we introduce a novel prototype alignment regularization by performing the few-shot segmentation in a reverse direction. Namely, the query image together with its predicted mask is considered as a new support set and used to segment the previous support images. In this way, the model is encouraged to generate more consistent prototypes between  support and query, offering better generalization performance.

Accordingly, we develop a Prototype Alignment Network (PANet) to tackle few-shot segmentation, as shown in Figure~\ref{fig:overall}. PANet first embeds different foreground objects and background into different prototypes via a shared feature extractor. In this way, each learned prototype is representative for the corresponding class and meanwhile is sufficiently distinguishable from other classes. Then, each pixel of the query image is labeled by referring to the class-specific prototypes nearest to its embedding representation. We find that even with only one support image per class, PANet can provide satisfactory segmentation results, outperforming the state-of-the-arts. Furthermore, it imposes a prototype alignment regularization by forming a new support set with the query image and its predicted mask and performing segmentation on the original support set. We find this indeed encourages the prototypes generated from the queries to align well with those of the supports. Note that the model is regularized only in training and the query images should be not confused with the testing images.

The structure design of the proposed PANet has several advantages. First, it introduces no extra learnable parameters and thus is less prone to over-fitting. Second, within PANet, the prototype embedding and prediction are performed on the computed feature maps and therefore segmentation requires no extra passes through the network. In addition, as the regularization is only imposed in training, the computation cost for inference does not increase.

Our few-shot segmentation model is a generic one. Any network with a fully convolutional structure can be used as the feature extractor. It also learns well from weaker annotations, \eg, bounding boxes or scribbles, as shown in experiments.  To sum up, the contributions of this work are:
\vspace{-15pt}
\begin{itemize}
\setlength\itemsep{0em}
    \item We propose a simple yet effective PANet for few-shot segmentation. The model exploits metric learning over prototypes, which differs from most existing works that adopt a parametric classification architecture.
    \vspace{-2pt}
    \item We propose a novel prototype alignment regularization to fully exploit the support knowledge to improve the few-shot learning.
    \vspace{-2pt}
    \item Our model can be directly applied  to learning from a few examples with  weak annotations.
    \vspace{-2pt}
    \item Our PANet  achieves  mIoU  of 48.1\% and 55.7\% on PASCAL-5\textsuperscript{i} for  1-shot and 5-shot settings, outperforming state-of-the-arts by a  margin up to 8.6 \%.
\end{itemize}


\section{Related work}
\paragraph{Semantic segmentation} 
Semantic segmentation aims to classify each pixel of an image into a set of predefined semantic classes. Recent methods are mainly based on deep convolutional neural networks~\cite{long2015fully, lin2017refinenet, badrinarayanan2017segnet, zhao2017pyramid, chen2018deeplab}. For example, Long \etal~\cite{long2015fully} first adopted deep CNNs and proposed Fully Convolutional Network (FCN) which greatly improves segmentation performance. Dilated convolutions~\cite{yu2015multi, chen2018deeplab} are widely used to increase the receptive field without losing spatial resolution. In this work, we follow the structure of FCN to perform dense prediction and also adopt dilated convolutions to enjoy a larger receptive field. Compared to models trained with full supervision, our model can generalize to new categories with only a handful of annotated data.

\vspace{-10pt}
\paragraph{Few-shot learning} Few-shot learning targets at learning transferable knowledge across different tasks with only a few examples. Many methods have been proposed, such as methods based on metric learning~\cite{vinyals2016matching, snell2017prototypical}, learning the optimization process~\cite{ravi2016optimization, finn2017model} and applying graph-based methods~\cite{garcia2018fewshot, liu2018learning}. Vinyals \etal~\cite{vinyals2016matching} encoded input into deep neural features and performed weighted nearest neighbor matching to classify unlabelled data. Snell \etal~\cite{snell2017prototypical} proposed a Prototypical Network to represent each class with one feature vector (prototype). Sung \etal~\cite{sung2018learning} used a separate module to directly learn the relation between support features and query features. Our model follows the Prototypical Network~\cite{snell2017prototypical} and can be seen as an extension of it to dense prediction tasks, enjoying a simple design yet high performance.

\vspace{-10pt}
\paragraph{Few-shot segmentation} 
Few-shot segmentation is receiving increasing interest recently. Shaban \etal~\cite{shaban2017one} first proposed a model for few-shot segmentation using a conditioning branch to generate a set of parameters $\theta$ from the support set, which is then used to tune the segmentation process of the query set. Rakelly \etal~\cite{rakelly2018conditional} concatenated extracted support features with  query ones and used a decoder to generate segmentation results. Zhang \etal~\cite{zhang2018sg} used masked average
pooling to better extract foreground/background information from the support set. Hu \etal~\cite{Hu2018AttentionbasedMG} explored guiding at multiple stages of the networks. These methods typically adopt a parametric module, which fuses information extracted from the support set and generates segmentation. 

\begin{figure*}[t!]
\begin{center}
   \includegraphics[width=\linewidth]{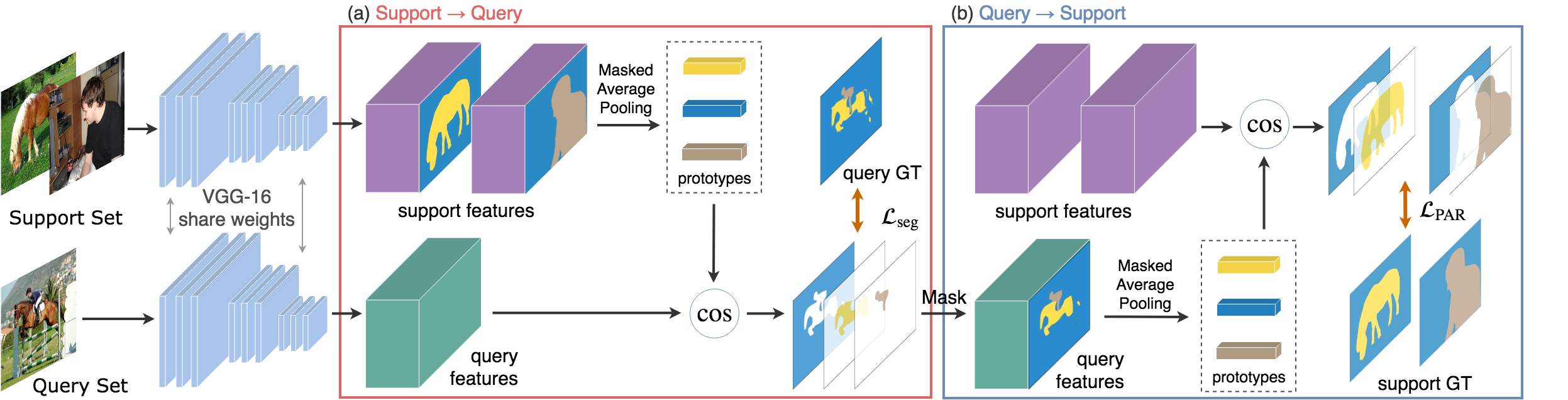}
\end{center}
   \caption{Illustration of the pipeline of our method in a 2-way 1-shot example. In block (a), PANet performs a support-to-query few-shot segmentation. The support and query images are embedded into deep features. Then the prototypes are obtained by masked average pooling. The query image is segmented via computing the cosine distance ($\cos$ in the figure) between each prototype and query features at each spatial location. Loss $\mathcal{L}_{\text{seg}}$ is computed between the segmentation result and the ground truth mask. In block (b), the proposed PAR aligns the prototypes of support and query by performing a query-to-support few-shot segmentation and calculating loss $\mathcal{L}_{\text{PAR}}$. GT denotes the ground truth segmentation masks.}
\label{fig:model}
\end{figure*}

Dong \etal~\cite{dong2018few} also adopted the idea of prototypical networks and tackled few-shot segmentation using metric learning. However, the model is too complex, involving three training stages and complicated training configurations. Besides, their method extracts prototypes based on an image-level loss and uses prototypes as guidance to tune the segmentation of the query set rather than obtaining segmentation directly from metric learning. Comparatively, our model has a simpler design and is more similar to the Prototypical Network~\cite{snell2017prototypical}. Besides, we adopt late fusion~\cite{rakelly2018few} to incorporate the annotation masks, making it easier to generalize to cases with sparse or updating annotations.

\section{Method}

\subsection{Problem setting}
We aim at obtaining a segmentation model that can learn fast to perform segmentation from only a few annotated images over new images from the same classes. As in previous works~\cite{shaban2017one},
we adopt the following model training and testing protocols. Suppose we are provided with images from two non-overlapping sets of classes $\mathcal{C}_{\text{seen}}$ and $\mathcal{C}_{\text{unseen}}$. The training set $\mathcal{D}_{\text{train}}$ is constructed from $\mathcal{C}_{\text{seen}}$ and the test set $\mathcal{D}_{\text{test}}$ is constructed from $\mathcal{C}_{\text{unseen}}$. We train the segmentation model $\mathcal{M}$  on $\mathcal{D}_{\text{train}}$ and evaluate on $\mathcal{D}_{\text{test}}$.

Both the training set $\mathcal{D}_{\text{train}}$ and testing set $\mathcal{D}_{\text{test}}$ consist of several \textit{episodes}.  Each episode is composed of a set of support images  $\mathcal{S}$ (with annotations) and a set of query images $\mathcal{Q}$.  Namely, $\mathcal{D}_{\text{train}}=\{(\mathcal{S}_i, \mathcal{Q}_i)\}_{i=1}^{N_{\text{train}}}$ and $\mathcal{D}_{\text{test}}=\{(\mathcal{S}_i, \mathcal{Q}_i)\}_{i=1}^{N_{\text{test}}}$, where $N_{\text{train}}$ and $N_{\text{test}}$ denote the number of episodes for training and testing respectively. 

Each training/testing episode $(\mathcal{S}_i, \mathcal{Q}_i)$ instantiates a $C$-way $K$-shot segmentation learning task. Specifically, the support set $\mathcal{S}_i$ has $K$ $\langle$image, mask$\rangle$ pairs per semantic class and there are in total $C$ different classes from $\mathcal{C_{\text{seen}}}$ for training and from $\mathcal{C_{\text{unseen}}}$ for testing, \ie $\mathcal{S}_i=\{(I_{c,k}, M_{c,k})\}$ where $k=1,2,\cdots,K$ and $c\in\mathcal{C}_i$ with $|\mathcal{C}_i|=C$.
The query set $\mathcal{Q}_i$ contains $N_{\text{query}}$ $\langle$image, mask$\rangle$ pairs from the same set of classes $\mathcal{C}_i$ as the support set. The model first extracts knowledge about the $C$ classes from the support set and then applies the learned knowledge to perform segmentation on the query set. As each episode contains different semantic classes, the model is trained to generalize well.
After obtaining the segmentation model $\mathcal{M}$ from the training set $\mathcal{D_{\text{train}}}$, we evaluate its few-shot segmentation performance on the test set $\mathcal{D}_{\text{test}}$ across all the episodes. In particular, for each testing episode the segmentation model $\mathcal{M}$ is evaluated on the query set $\mathcal{Q}_i$ given the support set $\mathcal{S}_i$.

\subsection{Method overview}

Different from existing few-shot segmentation methods which fuse the extracted support features with the query features to generate the segmentation results in a parametric way, our proposed model aims to learn and align compact and robust prototype representations for each semantic class in an embedding space. Then it performs segmentation within the embedding space via non-parametric metric learning.

As shown in Figure~\ref{fig:model}, our model learns to perform segmentation as follows. For each episode, it first embeds the support and query images into deep features by a shared backbone network. Then it applies the masked average pooling  to obtain prototypes from the support set, as detailed in Section~\ref{sec:method-prototype}. Segmentation over the query images is performed by labeling each pixel as the class of the nearest prototype. A novel prototype alignment regularization (PAR) introduced in Section~\ref{sec:method-par} is applied over the learning procedure to encourage the model to learn consistent embedding prototypes for the support and query.

We adopt a VGG-16~\cite{simonyan2014very} network as the feature extractor following conventions. The first 5 convolutional blocks in VGG-16 are kept for feature extraction and other layers are removed. The stride of \textit{maxpool4} layer is set to 1 for maintaining large spatial resolution. To increase the receptive field, the convolutions in \textit{conv5} block are replaced by dilated convolutions with dilation set to 2. As the proposed PAR introduces no extra learnable parameters, our network is trained end-to-end to optimize the weights of VGG-16 for learning a consistent embedding space.

\subsection{Prototype learning}  \label{sec:method-prototype}

Our model learns representative and well-separated prototype representation for each semantic class, including the background, based on the prototypical network~\cite{snell2017prototypical}. Instead of averaging over the whole input image~\cite{snell2017prototypical}, PANet leverages the mask annotations over the support images to learn prototypes for foreground and background separately. There are two strategies to exploit the segmentation masks \ie, early fusion and late fusion~\cite{rakelly2018few}. Early fusion masks the support images before feeding them into the feature extractor~\cite{shaban2017one,Hu2018AttentionbasedMG,dong2018few}. Late fusion directly masks over the feature maps to produce foreground/background features separately~\cite{zhang2018sg, rakelly2018conditional}. In this work, we adopt the late fusion strategy since it keeps the input consistency for the shared feature extractor. Concretely, given a support set $\mathcal{S}_i=\{(I_{c,k}, M_{c,k})\}$, let $F_{c,k}$ be the feature map output by the network for the image $I_{c,k}$. Here $c$ indexes the class and $k=1,\ldots,K$ indexes the support image. The prototype of class $c$ is computed via masked average pooling~\cite{zhang2018sg}:
\begin{equation}
\label{eqn:prototype_fg}
    p_c = \cfrac{1}{K} \sum_{k} \cfrac{\sum_{x,y} F_{c,k}^{(x,y)} \mathds{1}[M_{c,k}^{(x,y)}=c]}{\sum_{x,y} \mathds{1}[M_{c,k}^{(x,y)}=c]},
\end{equation}
where  $(x,y)$ indexes the spatial locations and $\mathds{1}(\cdot)$ is an indicator function, outputting value $1$ if the argument is true or $0$ otherwise.
In addition, the prototype of background is computed by
\begin{equation}
\label{eqn:prototype_bg}
    p_{\text{bg}} = \cfrac{1}{CK}\sum_{c,k} \cfrac{\sum_{x,y} F_{c,k}^{(x,y)} \mathds{1}[M_{c,k}^{(x,y)}\notin\mathcal{C}_i]}{\sum_{x,y} \mathds{1}[M_{c,k}^{(x,y)}\notin\mathcal{C}_i]}.
\end{equation}
The above prototypes are optimized end-to-end through non-parametric metric learning as explained below.

\subsection{Non-parametric metric learning}\label{dist}
We adopt a non-parametric metric learning method to learn the optimal prototypes and perform segmentation accordingly. Since segmentation can be seen as classification at each spatial location, we calculate the distance between the query feature vector at each spatial location with each computed prototype. Then we apply a softmax over the distances to produce a probability map $\tilde{M}_{q}$ over semantic classes (including background). Concretely, given a distance function $d$, let $\mathcal{P}=\{p_c|c\in\mathcal{C}_i\}\cup\{p_{\text{bg}}\}$ and $F_{q}$ denote the query feature map. For each $p_j\in\mathcal{P}$ we have
\begin{equation}
\label{eqn:prob_map}
    \tilde{M}_{q;j}^{(x,y)} = \cfrac{\exp(-\alpha d(F_{q}^{(x, y)}, p_j))}{\sum_{p_j\in\mathcal{P}}\exp(-\alpha d(F_{q}^{(x, y)}, p_j))}.
\end{equation}
The predicted segmentation mask is then given by
\begin{equation}
\label{eqn:seg_mask}
    \hat{M}_{q}^{(x,y)} = \argmax_j \tilde{M}_{q;j}^{(x,y)}.
\end{equation}

 The distance function $d$ commonly adopts the cosine distance or squared Euclidean distance. Snell \etal~\cite{snell2017prototypical} claimed using squared Euclidean distance greatly outperforms using cosine distance. However, Oreshkin \etal~\cite{oreshkin2018tadam} attributed the improvement to interaction of the different scaling of the metrics with the softmax function. Multiplying the cosine distance by a factor $\alpha$ can achieve comparable performance as using squared Euclidean distance. Empirically, we find that using cosine distance is more stable and gives better performance, possibly because it is bounded and thus easier to optimize. The multiplier $\alpha$ is fixed at 20 since we find learning it yields little performance gain.

After computing the probability map $\tilde{M}_{q}$ for the query image via metric learning, we calculate the segmentation loss $\mathcal{L}_{\text{seg}}$  as follows:
\begin{equation}
\label{eqn: loss_seg}
    \mathcal{L}_{\text{seg}} = -\cfrac{1}{N}\sum_{x,y}\sum_{p_j\in\mathcal{P}}\mathds{1}[M_{q}^{(x,y)}=j]\log  \tilde{M}_{q;j}^{(x,y)},
\end{equation}
where $M_{q}$ is the ground truth segmentation mask of the query image and $N$ is the total number of spatial locations. Optimizing the above loss will derive suitable prototypes for each class.

\subsection{Prototype alignment regularization (PAR)} \label{sec:method-par}

In previous works, the support annotations are used only for masking, which actually does not adequately exploit the support information for few-shot learning. In this subsection, we elaborate on the prototype alignment regularization (PAR) that exploits support information better to guide the few-shot learning procedure and helps enhance generalizability of the resulted model from a few examples. 

Intuitively, if the model can predict a good segmentation mask for the query using prototypes extracted from the support, the prototypes learned from the query set based on the predicted masks should be able to segment support images well. Thus, PAR encourages the resulted segmentation model to perform few-shot learning in a reverse direction, \ie, taking the query and the predicted mask as the new support to learn to segment the support images. This imposes a mutual alignment between the prototypes of support and query images and learns richer knowledge from the support. Note all the support and query images here are from the training set $\mathcal{D}_{\text{train}}$.

Figure~\ref{fig:model} illustrates PAR in details. After obtaining a segmentation prediction for the query image, we perform masked average pooling accordingly on the query features and obtain  another set of prototypes $\bar{\mathcal{P}}=\{\bar{p}_c|c\in\mathcal{C}_i\}\cup\{\bar{p}_{\text{bg}}\}$, following Eqns.~\eqref{eqn:prototype_fg} and~\eqref{eqn:prototype_bg}. Next, the non-parametric method introduced in Section~\ref{dist} is used to predict the segmentation masks for the support images. The predictions are compared with the ground truth annotations to calculate a loss $\mathcal{L}_{\text{PAR}}$. The entire procedure for implementing  PAR can be seen as swapping the support and query set. Concretely, within PAR, the segmentation probability of the support image $I_{c,k}$ is given by
\begin{equation}
\label{eqn:prob_map_supp}
    \tilde{M}_{c,k;j}^{(x,y)} = \cfrac{\exp(-\alpha d(F_{c,k}^{(x, y)}, \bar{p}_j))}{\sum_{\bar{p}_j\in \{\bar{p}_c, \bar{p}_{\text{bg}}\}}\exp(-\alpha d(F_{c,k}^{(x, y)}, \bar{p}_j))},
\end{equation}
and the loss $\mathcal{L}_{\text{PAR}}$ is computed by
\begin{equation}
\label{eqn:loss_par}
    \mathcal{L}_{\text{PAR}} = -\cfrac{1}{CKN}\sum_{c,k,x,y} \sum_{p_j\in\mathcal{P}}\mathds{1}[M_{q}^{(x,y)}=j]\log  \tilde{M}_{q;j}^{(x,y)}.
\end{equation}
Without PAR, the information only flows one-way from the support set to the query set. By flowing the information back to the support set, we force the model to learn a consistent embedding space that aligns the query and support prototypes. The aligning effect of the proposed PAR is validated by experiments in Section~\ref{experiment-par}.

The total loss for training our PANet model is thus
\begin{equation*}
    \mathcal{L} = \mathcal{L}_{\text{seg}} + \lambda \mathcal{L}_{\text{PAR}}. 
\end{equation*}
where $\lambda$ serves as regularization strength and $\lambda=0$ reduces to the model without PAR. In our experiments, we keep $\lambda$ as $1$ since different values give little improvement. The whole training and testing procedures for PANet on few-shot segmentation are summarized in Algorithm~\ref{algo}.

\subsection{Generalization to weaker annotations}

\begin{table*}[t!]
\centering
 \begin{tabular}{l|@{\hskip4.5pt}c@{\hskip4.5pt} @{\hskip4.5pt}c@{\hskip4.5pt} @{\hskip4.5pt}c@{\hskip4.5pt} @{\hskip4.5pt}c@{\hskip4.5pt} @{\hskip4.5pt}c@{\hskip4.5pt}| @{\hskip4.5pt}c@{\hskip4.5pt} @{\hskip4.5pt}c@{\hskip4.5pt} @{\hskip4.5pt}c@{\hskip4.5pt} @{\hskip4.5pt}c@{\hskip4.5pt} @{\hskip4.5pt}c@{\hskip4.5pt}|c|c} 
 \toprule
 \multirow{2}{*}{Method} &
 \multicolumn{5}{@{\hskip4.5pt}c|@{\hskip4.5pt}}{1-shot} &
 \multicolumn{5}{@{\hskip4.5pt}c@{\hskip4.5pt}|}{5-shot}
 & $\Delta$
 & \multirow{2}{*}{\#Params}
 \\
 \cmidrule{2-12}
 & split-1 & split-2
 & split-3 & split-4
 & Mean
 & split-1 & split-2
 & split-3 & split-4
 & Mean
 & Mean
 &
 \\
 \midrule
 OSLSM~\cite{shaban2017one}
        & 33.6 & 55.3 & 40.9 & 33.5 & 40.8 
        & 35.9 & 58.1 & 42.7 & 39.1 & 43.9
        & 3.1 & 272.6M
 \\
 co-FCN~\cite{rakelly2018conditional}$\dagger$
        & 36.7 & 50.6 & 44.9 & 32.4 &  41.1
        & 37.5 & 50.0 & 44.1 & 33.9 & 41.4
        & 0.3 & 34.2M
 \\
 SG-One~\cite{zhang2018sg}
        & 40.2 & \textbf{58.4} & 48.4 & 38.4 & 46.3
        & 41.9 & 58.6 & 48.6 & 39.4 & 47.1
        & 0.8 & 19.0M
 \\
 PANet-init
        & 30.8 & 40.7 & 38.3 & 31.4 & 35.3 
        & 41.6 & 52.7 & 51.6 & 40.8 & 46.7
        & \textbf{11.4} & 14.7M
 \\
 PANet
        & \textbf{42.3} & 58.0 & \textbf{51.1} & \textbf{41.2} & \textbf{48.1}
        & \textbf{51.8} & \textbf{64.6} & \textbf{59.8} & \textbf{46.5} & \textbf{55.7}
        & 7.6 & \textbf{14.7M}
 \\
 \bottomrule
 \end{tabular}
 \caption{Results of 1-way 1-shot and 1-way 5-shot segmentation on PASCAL-5\textsuperscript{i} dataset using mean-IoU metric. $\Delta$ denotes the difference between 1-shot and 5-shot. $\dagger$: The results of co-FCN in mean-IoU metric are reported by~\cite{zhang2018sg}.}
\label{table:pascal_result}
\end{table*}

Our model is generic and is directly applicable to other types of annotations. First, it accepts weaker annotations on the support set, such as scribbles and bounding boxes indicating the foreground objects of interest. Experiments in Section~\ref{sec:weak} show that even with weak annotations, our model is still able to extract robust prototypes from the support set and give comparably good segmentation results for the query images. Compared with pixel-level dense annotations, weak annotations are easier and cheaper to obtain~\cite{lin2016scribblesup}. Second, by adopting late fusion~\cite{rakelly2018few}, our model can quickly adapt to updated annotations with little computation overhead  and thus can be applied in interactive segmentation. We leave this for future works.

\begin{algorithm}[t]
\SetAlgoLined
\SetAlgoHangIndent{12pt}
\SetKwInOut{Input}{Input}\SetKwInOut{Output}{Output}
\Input{A training set $\mathcal{D}_{\text{train}}$ and a testing set $\mathcal{D}_{\text{test}}$}
\For{\text{each episode} $(\mathcal{S}_i, \mathcal{Q}_i)\in \mathcal{D}_{\text{train}}$}
{
Extract prototypes $\mathcal{P}$ from the support set $\mathcal{S}_i$ using Eqns.~\eqref{eqn:prototype_fg} and~\eqref{eqn:prototype_bg}\\
Predict the segmentation probabilities and masks for the query image using Eqns.~\eqref{eqn:prob_map} and~\eqref{eqn:seg_mask}\\
Compute the loss $\mathcal{L}_{\text{seg}}$ as in Eqn.~\eqref{eqn: loss_seg} \\
Extract prototypes $\bar{\mathcal{P}}$ from the query set $\mathcal{Q}_i$ using Eqns.~\eqref{eqn:prototype_fg} and~\eqref{eqn:prototype_bg} \\
Predict segmentation probabilities for the support images using Eqn.~\eqref{eqn:prob_map_supp} \\
Compute the loss $\mathcal{L}_{\text{PAR}}$ as in Eqn.~\eqref{eqn:loss_par}\\
Compute the gradient and optimize via SGD
}

\For{\text{each episode} $(\mathcal{S}_i, \mathcal{Q}_i)\in \mathcal{D}_{\text{test}}$}
{
Extract prototypes $\mathcal{P}$ from the support set $\mathcal{S}_i$ using Eqns.~\eqref{eqn:prototype_fg} and~\eqref{eqn:prototype_bg}\\
Predict the segmentation probabilities and masks for the query image using Eqns.~\eqref{eqn:prob_map} and~\eqref{eqn:seg_mask}
}

  \caption{Training and evaluating PANet.}\label{algo}
\end{algorithm}

\section{Experiments}

\subsection{Setup}
 
 \paragraph{Datasets} We follow the evaluation scheme proposed in~\cite{shaban2017one} and evaluate our model on the  {PASCAL-5\textsuperscript{i}}~\cite{shaban2017one} dataset. The dataset is created from PASCAL VOC 2012~\cite{everingham2010pascal} with SBD~\cite{hariharan2011semantic} augmentation. The 20 categories in PASCAL VOC are evenly divided into 4 splits, each containing 5 categories.  Models are trained on 3 splits and evaluated on the rest one in a cross-validation fashion. The categories in each split can be found in~\cite{shaban2017one}. During testing, previous methods randomly sample 1,000 episodes for evaluation but we find it is not enough to give stable results. In our experiments, we average the results from 5 runs with different random seeds, each run containing 1,000 episodes.
 
 Following~\cite{Hu2018AttentionbasedMG}, we also evaluate our model on a more challenging dataset built from MS COCO~\cite{lin2014microsoft}. Similarly, the 80 object classes in MS COCO are evenly divided into 4 splits, each containing 20 classes. We follow the same scheme for training and testing as on the PASCAL-5\textsuperscript{i}. $N_{\text{query}}=1$ is used for all experiments.

\begin{table}[t!]
\centering
 \begin{tabular}{l|c c c} 
 \toprule
 Method & 1-shot & 5-shot & $\Delta$
 \\
 \midrule
 FG-BG~\cite{rakelly2018conditional}
        & 55.0 & - & -
 \\
 Fine-tuning~\cite{rakelly2018conditional}
        & 55.1 & 55.6 & 0.5
 \\
 OSLSM~\cite{shaban2017one}
        & 61.3 & 61.5 & 0.2
 \\
 co-FCN~\cite{rakelly2018conditional}
        & 60.1 & 60.2 & 0.1
 \\
 PL~\cite{dong2018few}
        & 61.2 & 62.3 & 1.1
 \\
 A-MCG~\cite{Hu2018AttentionbasedMG}
        & 61.2 & 62.2 & 1.0
 \\
 SG-One~\cite{zhang2018sg}
        & 63.9 & 65.9 & 2.0
 \\
 PANet-init
        & 58.9 & 65.7 & \textbf{6.8}
 \\
 PANet
        & \textbf{66.5} & \textbf{70.7} & 4.2
 \\
 \bottomrule
 \end{tabular}
 \caption{Results of 1-way 1-shot and 1-way 5-shot segmentation on PASCAL-5\textsuperscript{i} dataset using binary-IoU metric. $\Delta$ denotes the difference between 1-shot and 5-shot.}
\label{table:pascal_result_binaryIoU}
\end{table}

\vspace{-12pt}
\paragraph{Evaluation metrics} We adopt two metrics for model evaluation, mean-IoU and binary-IoU. Mean-IoU measures the Intersection-over-Union (IoU) for each foreground class and averages over all the classes~\cite{shaban2017one, zhang2018sg}. Binary-IoU treats all object categories as one foreground class and averages the IoU of foreground and background~\cite{rakelly2018conditional, dong2018few, Hu2018AttentionbasedMG}. We mainly use the mean-IoU metric because it considers the differences between foreground categories and therefore more accurately reflects  the model performance. Results w.r.t.\ the binary-IoU are also reported for clear comparisons with some previous methods.

\vspace{-12pt}
\paragraph{Implementation details} We initialize the VGG-16 network with the weights pre-trained on ILSVRC~\cite{ILSVRC15} as in previous works~\cite{shaban2017one, dong2018few, zhang2018sg}. Input images are resized to (417, 417) and augmented using random horizontal flipping. The model is trained end-to-end by SGD with the momentum of 0.9 for 30,000 iterations. The learning rate is initialized to 1e-3 and reduced by 0.1 every 10,000 iterations. The weight decay is 0.0005 and the batch size is 1.

\vspace{-11pt}
\paragraph{Baselines} We set a baseline model which is initialized with the weights pre-trained on ILSVRC~\cite{ILSVRC15} but not further trained on PASCAL-5\textsuperscript{i}, denoted as PANet-init. We also compare our PANet with two baseline models FG-BG and fine-tuning from~\cite{rakelly2018conditional}. FG-BG trains a foreground-background segmentor which is independent of the support and fine-tuning is used to tune a pre-trained foreground-background segmentor on the support.

\subsection{Comparison with state-of-the-arts}

\paragraph{PASCAL-5\textsuperscript{i}} Table~\ref{table:pascal_result} compares our model with other methods on PASCAL-5\textsuperscript{i} dataset in mean-IoU metric. Our model outperforms the state-of-the-art methods in both 1-shot and 5-shot settings while using fewer parameters. In the 5-shot task, our model achieves significant improvement of 8.6\%.  Using binary-IoU metric, as shown in Table~\ref{table:pascal_result_binaryIoU}, our model also achieves the highest performance. It is worth noting that our method does not use any decoder module or post-processing techniques to refine the results.

\begin{table}[t!]
\centering
 \begin{tabular}{l|c c|c c} 
 \toprule
 \multirow{2}{*}{Method} &
 \multicolumn{2}{c|}{mean-IoU} &
 \multicolumn{2}{c}{binary-IoU}
 \\
 \cmidrule{2-5}
 & 1-shot & 5-shot & 1-shot & 5-shot
 \\
 \midrule
 PL~\cite{dong2018few}
        & - & - & 42.7 & 43.7
 \\
 PANet
        & \textbf{45.1} & \textbf{53.1} & \textbf{64.2} & \textbf{67.9}
 \\
 \bottomrule
 \end{tabular}
 \caption{Results of 2-way 1-shot and 2-way 5-shot segmentation on PASCAL-5\textsuperscript{i} dataset.}
\label{table:pascal_result_multiway}
\end{table}

\begin{table}[t!]
\centering
 \begin{tabular}{l|cc|cc} 
 \toprule
 \multirow{2}{*}{Method} &
 \multicolumn{2}{c|}{mean-IoU} &
 \multicolumn{2}{c}{binary-IoU}
 \\
 \cmidrule{2-5}
 & 1-shot & 5-shot & 1-shot & 5-shot
 \\
 \midrule
 A-MCG~\cite{Hu2018AttentionbasedMG}
        & - & - & 52 & 54.7
 \\
 PANet
        & \textbf{20.9} & \textbf{29.7} & \textbf{59.2} & \textbf{63.5}
 \\
 \bottomrule
 \end{tabular}
 \caption{Results of 1-way 1-shot and 1-way 5-shot segmentation on MS COCO dataset.}
\label{table:coco_result}
\end{table}

\begin{figure*}[t!]
\begin{center}
   \includegraphics[width=\linewidth]{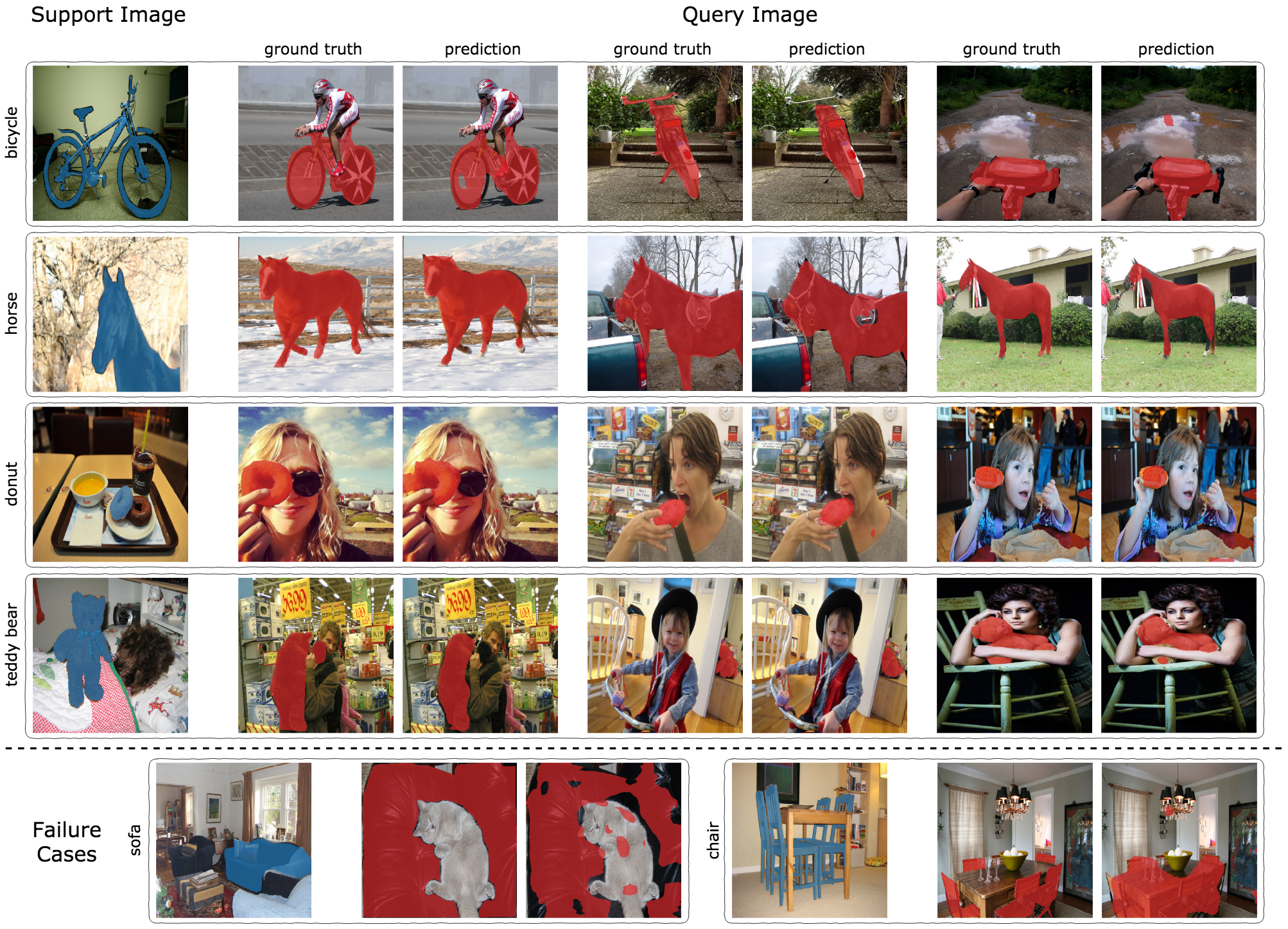}
\end{center}
   \caption{Qualitative results of our model in 1-way 1-shot segmentation on PASCAL-5\textsuperscript{i} (row 1 and 2) and MS COCO (row 3 and 4).}
\label{fig:1way1shot}
\end{figure*}

As Tables~\ref{table:pascal_result} and~\ref{table:pascal_result_binaryIoU} show, the performance gap between 1-shot and 5-shot settings is small in other methods (less than 3.1\% in mean-IoU), implying these methods obtain little improvement with more support information. In contrast, our model yields much more significant performance gain (up to 7.6\% in mean-IoU) since it learns more effectively from the support set. The evaluation results of our baseline model PANet-init also confirm this point. Without training, it rivals the state-of-the-art in 5-shot settings and gains more than 11\% in mean-IoU when given more support images.

As in~\cite{dong2018few, zhang2018sg}, we evaluate our model on multi-way few-shot segmentation tasks. Without loss of  generality, we perform evaluations on 2-way 1-shot and 2-way 5-shot segmentation tasks. Table~\ref{table:pascal_result_multiway} summarizes the results. Our PANet outperforms previous works by a large margin of more than 20\%.

\begin{figure}[t!]
\begin{center}
   \includegraphics[width=\linewidth]{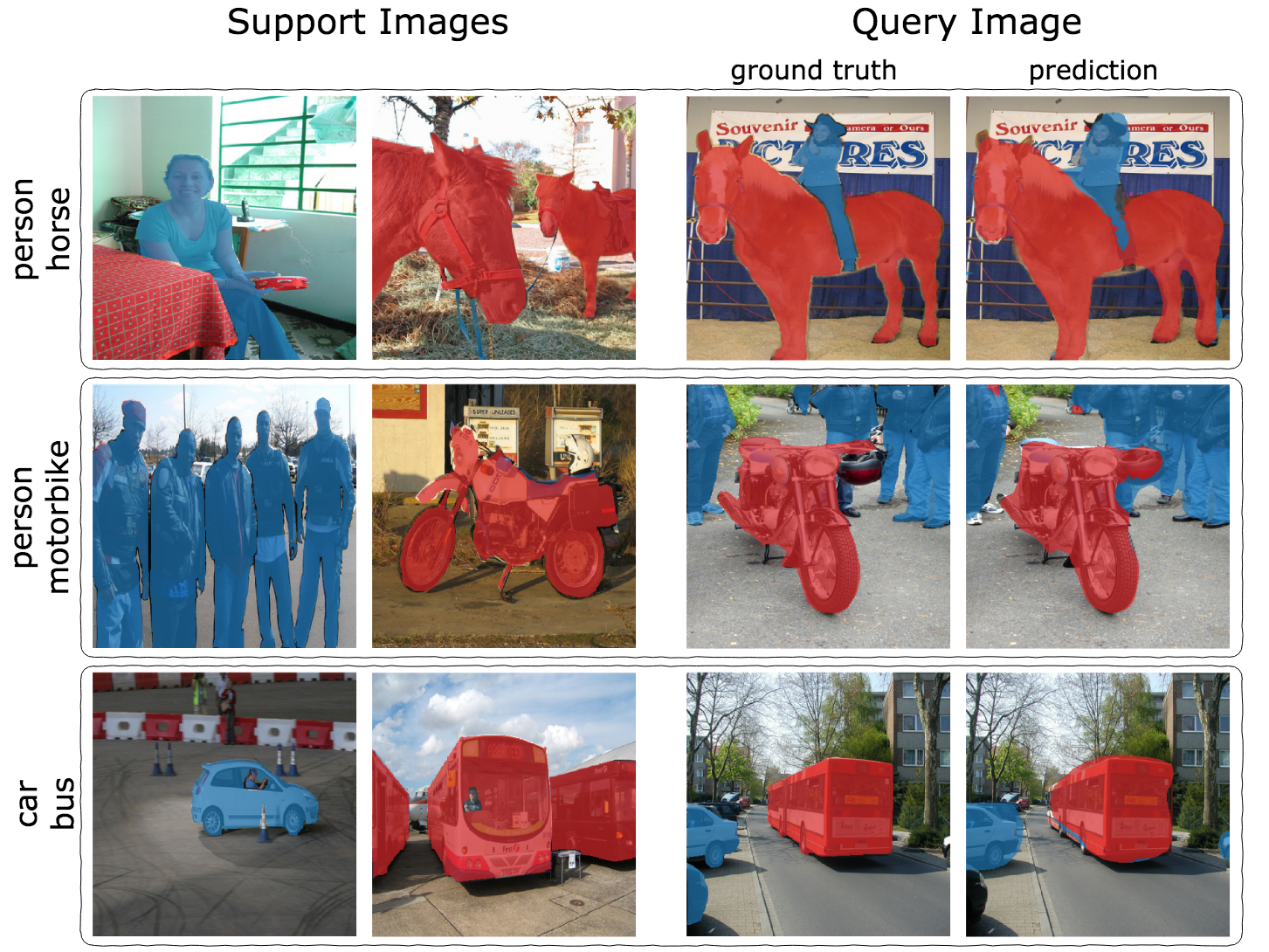}
\end{center}
   \caption{Qualitative results of our model in 2-way 1-shot segmentation on PASCAL-5\textsuperscript{i}.}
\label{fig:2way1shot}
\end{figure}

Qualitative results for 1-way and 2-way segmentation are shown in Figure~\ref{fig:1way1shot} and Figure~\ref{fig:2way1shot}. Without any decoder structure or post-processing, our model gives satisfying segmentation results on unseen classes with only one annotated support image. This demonstrates the strong learning  and generalization abilities of our model. Note that the prototype extracted from the same support image can be used to successfully segment the query images with appearance variations. For example, in Figure~\ref{fig:1way1shot} row 1, our model successfully segments bicycles: cluttered with other objects (1st example), viewed from a different perspective (2nd example), with only parts shown (3rd example). On the other hand, prototypes extracted from one part of the object can be used to segment whole objects of the same class (row 2 in Figure~\ref{fig:1way1shot}). It demonstrates that the proposed PANet is capable of extracting robust prototypes for each semantic class from a few annotated data. More qualitative examples can be found in the supplementary material.

We also present some challenging cases that fail our model. As the first failure case in Figure~\ref{fig:1way1shot} shows, our model tends to give segmentation results with unnatural patches, possibly because it predicts independently at each location. But this can be alleviated by post-processing. From the second failure case, we find our model is unable to distinguish between chairs and tables since they have similar prototypes in the embedding space.

\vspace{-10pt}
 \paragraph{MS COCO} Table~\ref{table:coco_result} shows the evaluation results on MS COCO dataset. Our model outperforms the previous A-MCG~\cite{Hu2018AttentionbasedMG}  by 7.2\% in 1-shot setting and 8.2\% in 5-shot setting. Compared to PASCAL VOC, MS COCO has more object categories, making the differences between two evaluation metrics more significant. Qualitative results on MS COCO are shown in Figure~\ref{fig:1way1shot}.

\vspace{-2mm}
\subsection{Analysis on PAR} \label{experiment-par}

The proposed PAR encourages the model to learn a consistent embedding space which aligns the support and query prototypes. Apart from minimizing the distances between the support and query prototypes, the models trained with PAR get better results (shown in Table~\ref{table:ablation}) as well as faster convergence of the training process.

\vspace{-12pt}
\paragraph{Aligning embedding prototypes} By flowing the information from the query set back to the support set via PAR, our model can learn a consistent embedding space and align the prototypes extracted from the support and query set. To verify this, we randomly choose 1,000 episodes from PASCAL-5\textsuperscript{i} split-1 in the 1-way 5-shot task. Then for each episode we calculate the Euclidean distance between prototypes extracted from the query set and the support set. The averaged distance computed by models with PAR is 32.2, much smaller than 42.6 by models without PAR. With PAR, our model is able to extract prototypes that are better aligned in the embedding space.

\begin{table}[t]
\centering
 \begin{tabular}{l|cc} 
 \toprule
 Method & 1-shot & 5-shot
 \\
 \midrule
 PANet w/o  PAR & 47.2 & 54.9
 \\
 PANet & 48.1 & 55.7
 \\
 \bottomrule
 \end{tabular}
 \caption{Evaluation results of our PANet trained with and without PAR on PASCAL-5\textsuperscript{i} in mean-IoU metric.}
\label{table:ablation}
\end{table}

\begin{table}[t!]
\centering
 \begin{tabular}{c |c c}
 \toprule
 Annotations & 1-shot & 5-shot
 \\
 \midrule
 Dense & 48.1 & 55.7
 \\
 Scribble & 44.8 & 54.6
 \\
 Bounding box & 45.1 & 52.8
 \\
 \bottomrule
 \end{tabular}
 \caption{Results of using different types of annotations in mean-IoU metric.}
\label{table:pascal_result_sparse}
\end{table}

\begin{figure}[t!]
\begin{center}
   \includegraphics[width=0.95\linewidth]{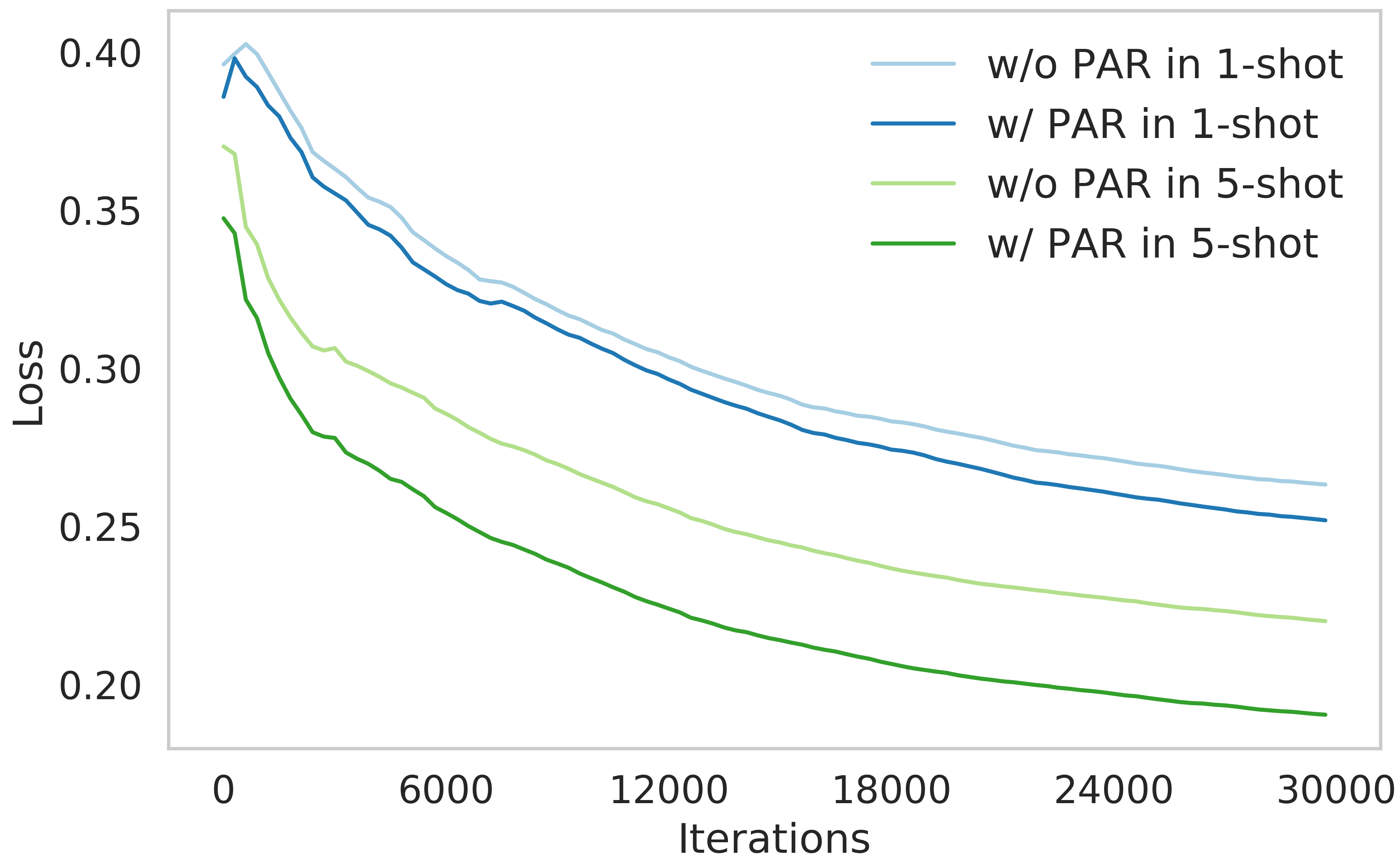}
\end{center}
   \caption{Training loss of models with and without PAR.}
\label{fig:converge}
\end{figure}

\vspace{-12pt}
\paragraph{Speeding up convergence} In our experiments, we observe that models trained  with  PAR  converge faster than models without it, as reflected from the training loss curve  in Figure~\ref{fig:converge}. This shows the PAR accelerates convergence and helps the model reach a lower loss, especially in 5-shot setting, because with PAR the information from the support set can be better exploited.

\subsection{Test with weak annotations} \label{sec:weak}

\begin{figure}[t!]
\begin{center}
   \includegraphics[width=0.95\linewidth]{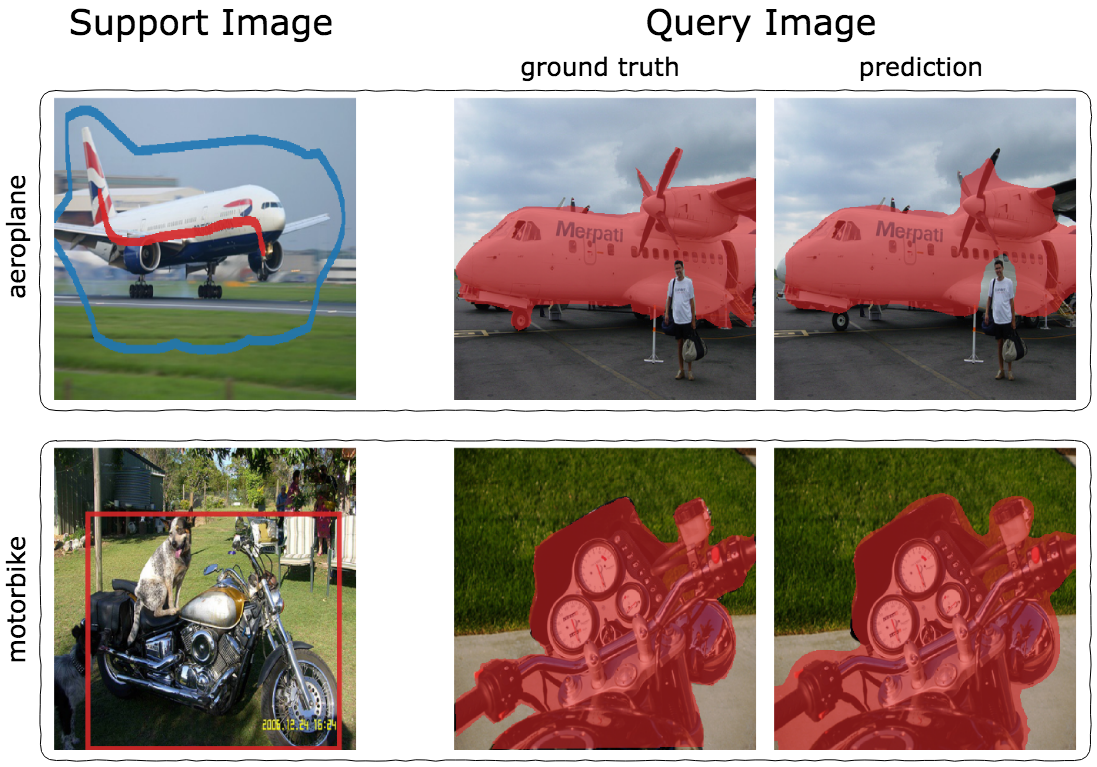}
\end{center}
   \caption{Qualitative results of our model on 1-way 1-shot segmentation using scribble and bounding box annotations. The scribbles are dilated for better visualization.}
\label{fig:sparse}
\end{figure}

We further evaluate our model with scribble and bounding box annotations. During testing, the pixel-level annotations of the support set are replaced by scribbles or bounding boxes which are generated from the dense segmentation masks automatically. Each bounding box is obtained from one randomly chosen instance mask in each support image. As Table~\ref{table:pascal_result_sparse} shows, our model works pretty well with very sparse annotations and is robust to the noise brought by the bounding box. In 1-shot learning case, the model performs comparably well with two different annotations,  but for 5-shot learning, using scribbles outperforms using bounding box by 2\%. A possible reason is with more support information, scribbles give more representative prototypes while bounding boxes introduce more noise.
Qualitative results of using scribble and bounding box annotations are shown in Figure~\ref{fig:sparse}. 

\section{Conclusion}
We propose a novel PANet for few-shot segmentation based on metric learning. PANet is able to extract robust prototypes from the support set and performs segmentation using non-parametric distance calculation. With the proposed PAR, our model can further exploit the support information to assist training. Without any decoder structure or post-processing step, our PANet outperforms previous work by a large margin.

\vspace{-5pt}
\paragraph{Acknowledgements} Jiashi Feng was partially supported by NUS IDS R-263-000-C67-646,  ECRA R-263-000-C87-133 and MOE Tier-II R-263-000-D17-112.

{\small
\bibliographystyle{ieee_fullname}
\bibliography{egbib}
}

\end{document}